\documentclass[journal,twoside]{IEEEtran}
\usepackage[ruled,vlined]{algorithm2e}
\usepackage{graphicx,amssymb,hyperref}
\usepackage{epstopdf}
\usepackage{booktabs}
\usepackage{color}
\usepackage{multirow}
\usepackage{amsmath}
\usepackage{subfigure}
\usepackage{bm}
\usepackage{url}
\usepackage{bbding}
\usepackage{indentfirst}
\usepackage{bbm}
\usepackage[T1]{fontenc}
\usepackage{cite}

\usepackage{tabu}
\usepackage[absolute]{textpos}
\newcommand{\copyrightstatement}{
    \begin{textblock}{15}(0.5,0.3)    
         \noindent
         \centering
         \textblockcolour{white}
         \footnotesize
         \copyright 2022 IEEE. Personal use of this material is permitted. Permission from IEEE must be obtained for all other uses, in any current or future media, including reprinting/republishing this material for advertising or promotional purposes, creating new collective works, for resale or redistribution to servers or lists, or reuse of any copyrighted component of this work in other works. DOI: 10.1109/TIP.2022.3188061
    \end{textblock}
}

\begin{document}

\title{MetaAge: Meta-Learning Personalized Age Estimators}

\copyrightstatement

\author{Wanhua~Li, Jiwen~Lu,~\IEEEmembership{Senior~Member,~IEEE}, Abudukelimu~Wuerkaixi,
Jianjiang~Feng, \IEEEmembership{Member,~IEEE}, and Jie~Zhou,~\IEEEmembership{Senior~Member,~IEEE}
\thanks{
This work was supported in part by the National Key Research and Development Program of China under Grant 2017YFA0700802, in part by the National Natural Science Foundation of China under Grant 62125603 and Grant U1813218, in part by a grant from the Beijing Academy of Artificial Intelligence (BAAI).
\emph{(Corresponding author: Jiwen Lu)}

Wanhua Li, Abudukelimu~Wuerkaixi, Jianjiang Feng, and Jie Zhou are with the Beijing National Research Center for Information Science and Technology (BNRist), and the Department of Automation, Tsinghua University, Beijing, 100084, China.
E-mail: li-wh17@mails.tsinghua.edu.cn;~wekxabdk17@mails.tsinghua.edu.cn;~jfeng@tsinghua.edu.cn;~jzhou@tsinghua.edu.cn.

Jiwen Lu is with the Beijing National Research Center for Information Science and Technology (BNRist), the Department of Automation, Tsinghua University, and the Beijing Academy of Artificial Intelligence, Beijing, 100084, China. E-mail: lujiwen@tsinghua.edu.cn.

}
\markboth{IEEE Transactions on Image Processing}%
{Li~\MakeLowercase{\textit{et al.}}: MetaAge: Meta-Learning Personalized Age Estimators}
}
\maketitle


\begin{abstract}
  Different people age in different ways. Learning a personalized age estimator for each person is a promising direction for age estimation given that it better models the personalization of aging processes. However, most existing personalized methods suffer from the lack of large-scale datasets due to the high-level requirements: identity labels and enough samples for each person to form a long-term aging pattern. In this paper, we aim to learn personalized age estimators without the above requirements and propose a meta-learning method named MetaAge for age estimation. Unlike most existing personalized methods that learn the parameters of a personalized estimator for each person in the training set, 
  our method learns the mapping from identity information to age estimator parameters.
  Specifically, we introduce a personalized estimator meta-learner, which takes identity features as the input and outputs the parameters of customized estimators. In this way, our method learns the meta knowledge without the above requirements and seamlessly transfers the learned meta knowledge to the test set, which enables us to leverage the existing large-scale age datasets without any additional annotations. Extensive experimental results on three benchmark datasets including MORPH II, ChaLearn LAP 2015 and ChaLearn LAP 2016 databases demonstrate that our MetaAge significantly boosts the performance of existing personalized methods and outperforms the state-of-the-art approaches.
\end{abstract}
\IEEEpeerreviewmaketitle

\section{Introduction}
\IEEEPARstart
In recent years, age prediction, as known as  age estimation has drawn a lot of attention in the computer vision community owing to its wide potential applications in surveillance monitoring \cite{dibekliouglu2015combining}, electronic customer relationship management \cite{fu2010age}, human-computer interaction (HCI) \cite{geng2006learning}, security control \cite{hu2016facial}, and biometrics \cite{lu2015cost}. Despite decades of efforts \cite{lanitis2002toward,guo2009human,chang2011ordinal,niu2016ordinal,chen2017using} have been devoted to age estimation, it remains a very challenging problem.

\begin{figure}[t]
  \centering
  \includegraphics[width=1.0\linewidth]{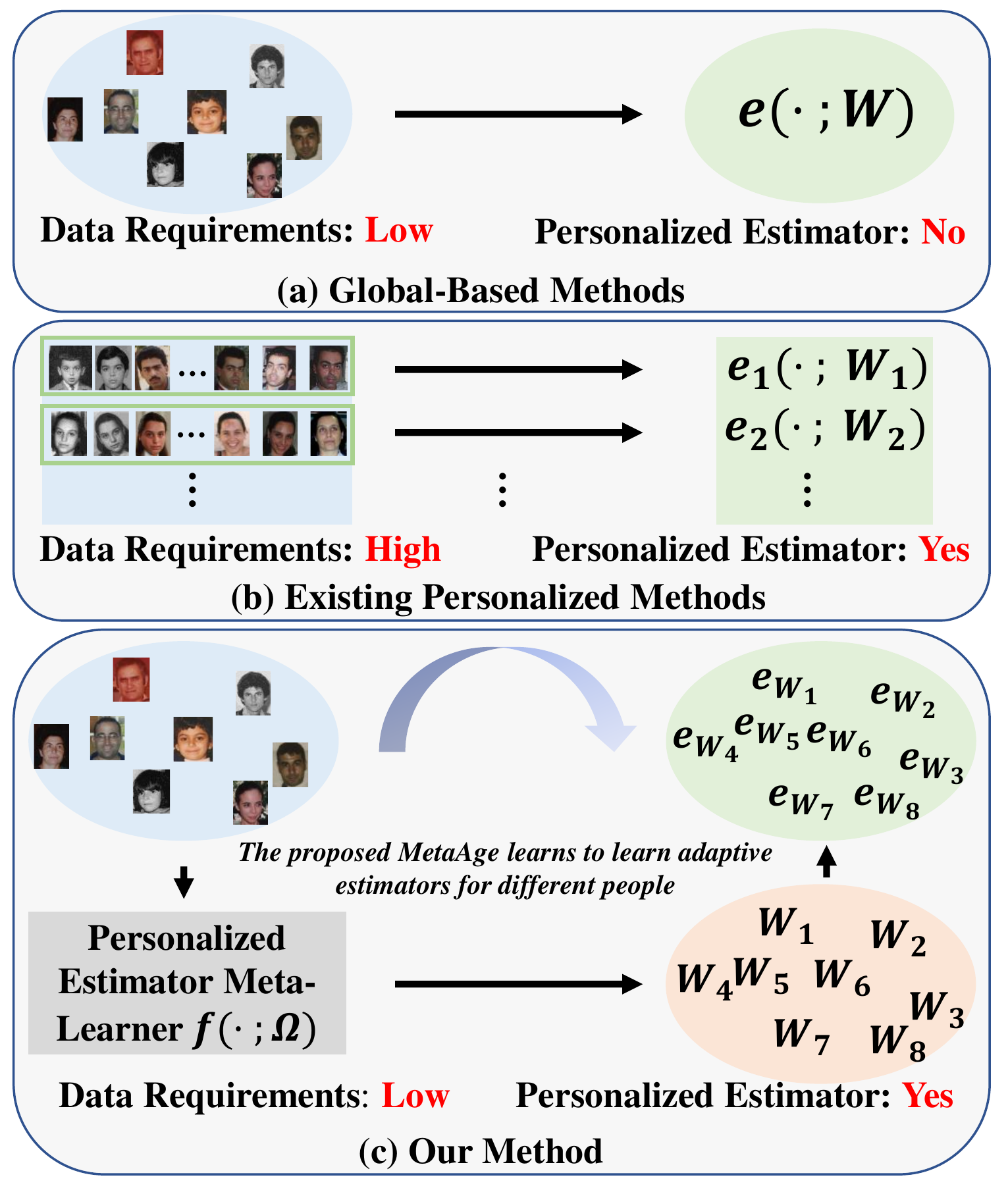}
  \caption{The key differences of global-based age estimation methods, existing personalized age estimation methods, and our method. Both $\bm{e(\cdot ; W)}$ and $\bm{e_{W}}$ are used to denote an age estimator parameterized by $\bm{W}$. Global-based age estimation methods only learn \emph{one global estimator} for all samples, whereas most existing personalized methods require that everyone in the training set has enough images and then train a \emph{personalized estimator} for each person. By contrast, our method learns to learn personalized estimators and outputs the parameters of an adaptive estimator for each person without the above two requirements.}
  \label{fig:introduction}
\end{figure}

One of the main challenges for facial age estimation is that different people age in different ways \cite{geng2007automatic}, \emph{i.e.}, different people go through different aging patterns. For example, different populations determined by intrinsic human genes, such as Asian and Caucasian, females and males, usually exhibit quite different aging patterns \cite{li2018deep}. Existing approaches for age estimation can be grouped into two categories \cite{zhang2010multi}: global-based age estimation methods and personalized age estimation methods. Global-based age estimation assumes that the aging processes are the same for different people and learns a global age estimator for all different people. On the other hand, personalized age estimation approaches learn personalized age estimators for different people. Personalized age estimation methods generally outperform global-based age estimation methods among all non-deep learning methods since they better model the unique characteristic of the aging processes \cite{zhang2010multi}.
Fig. \ref{fig:introduction}(a) and Fig. \ref{fig:introduction}(b) show the key differences between global-based age estimation methods and existing personalized age estimation methods.  Global-based age estimation methods utilize the existing age datasets without additional annotations. However, they only learn \emph{one global estimator} for all samples with different identities. 
Most existing personalized age estimation methods \cite{zhang2010multi,lanitis2002toward} usually learn the parameters of a person-specific age estimator for each person in the training set, which naturally brings two  high-level requirements to datasets:  \emph{identity labels} and \emph{enough images at different ages for each person} to form a long term aging pattern.

For personalized age estimation, there are no large-scale age datasets that meet either of the above two requirements.
Although we can hire human workers to label the identities, collecting a large-scale age dataset, where each person has images that cover a long-range age distribution, poses enormous challenges.
Meanwhile, global-based age estimation methods only require the dataset to be annotated with age labels, which is satisfied by any age dataset. With the rapid development of deep learning, global-based age estimation methods have made significant progress  \cite{pan2018mean,shen2018deep,li2019bridgenet} in recent years owing to the availability of large image repositories and high-performance computing systems. However, the lack of large-scale datasets that meet the above requirements makes existing personalized methods unable to effectively leverage the data-driven technologies of deep learning, which has become a major obstacle to the development of personalized age estimation methods.

To address the above two requirements, we propose a method named MetaAge to meta-learn personalized age estimators, which learns the mapping from identity information to age estimator parameters.
Although there are only a few samples for each person, the training set does contain many identities. Encouraged by the success of meta-learning in few-shot learning~\cite{finn2017model}, we consider \emph{learning to learn personalized estimators} rather than directly learning the parameters of estimators. Concretely, our MetaAge consists of a personalized estimator meta-learner, which takes identity features as the input and outputs the parameters of customized estimators. Our method can transfer the learned meta knowledge to any given unseen person since the identity features provide a unified semantic representation on the training set and test set. 
Fig. \ref{fig:introduction}(c) further shows the advantages of our method.  Compared with global-based age estimation methods and existing personalized age estimation methods,  our method learns to learn personalized age estimators without high-level requirements for age databases.

To summarize, the main contributions of this work are described as follows:

\begin{itemize}
\item
  To the best of our knowledge, the MetaAge is the first personalized age estimation method without the requirements of identity labels and enough samples for each person, which sheds light on data-driven personalized age estimation methods.
\item
  Different from existing methods that directly learn a personalized estimator for each person, our MetaAge proposes a personalized estimator meta-learner, which learns the mapping from identity information to age estimator parameters.
\item
  Experimental results on three benchmarks show that our approach not only largely improves the performance of personalized age estimation approaches but also outperforms state-of-the-art methods.

\end{itemize}

The remainder of this paper is organized as follows. We first give a brief review of the related work in Section II. Then we detail the proposed MetaAge in Section III. The experimental results and analysis are presented in Section IV. Finally, we conclude this paper in Section V.

\section{Related Work}
In this section, we briefly review two related topics including facial age estimation and meta learning.

\subsection{Facial Age Estimation}
Numerous facial age estimation methods \cite{lanitis2002toward,guo2010human,liu2019structure,he2017data} have been proposed over the past two decades, which can be mainly divided into two categories \cite{zhang2010multi}: global-based age estimation methods and personalized age estimation methods. Global-based age estimation methods usually learn a global age estimator for all different people, while personalized methods learn a personalized age estimator for each person.

Many of the early age estimation methods are global-based. For example, Guo \emph{et al.} \cite{guo2009human} first introduced the biologically inspired features (BIF) and achieved promising results.
Fu \emph{et al.} \cite{fu2008human} proposed a manifold learning approach to model the manifold representation with a multiple linear regression procedure based on a quadratic function. Xiao \emph{et al.} \cite{xiao2009learning} considered age estimation as a regression problem to learn a distance metric that measured the semantic similarity of the input data. Chang \emph{et al.} \cite{chang2011ordinal} presented the OHRank which formulated the age estimation problem as a series of sub-problems of binary classifications. Li \emph{et al.} \cite{li2012learning} further exploited the ordinal information among aging faces and presented a feature selection approach.

As personalized methods better model the unique characteristic of aging processes, they usually demonstrate more promising results. Geng \emph{et al.} \cite{geng2006learning,geng2007automatic} proposed a subspace approach called AGES to model the personalized aging patterns.
A multi-task extension of the warped Gaussian process was presented in \cite{zhang2010multi} by formulating age estimation as a multi-task learning problem where each task referred to the estimation of the age function of each person.
They defined an aging pattern as a sequence of personal face images sorted in time order and regarded each aging pattern as a sample instead of an isolated face image. The AGES utilized principal component analysis to find a representative linear subspace and estimated the age of a previously unseen face image by minimizing the reconstruction error.
A nonlinear extension was presented in \cite{geng2008facial} to learn a nonlinear aging pattern subspace. Geng \emph{et al.} \cite{geng2009facial} further assembled the face images in a higher-order tensor and developed a multilinear subspace analysis algorithm to learn both common features and person-specific  features automatically.

It is difficult to collect images of a person at different ages, so for any existing large-scale age dataset, the data of aging patterns are extremely insufficient. This limits the development of personalized methods. 
On the other hand, global-based methods have no such requirements for datasets. In recent years,  global-based methods have made significant  progress \cite{niu2016ordinal,shen2018deep,pan2018mean,shen2017label} due to the powerful feature representation of CNNs. Rothe \emph{et al.} \cite{rothe2018deep} posed age estimation as a deep classification problem and introduced the IMDB-WIKI dataset for pre-training. Niu \emph{et al.} \cite{niu2016ordinal} proposed a multiple output CNN to utilize the ordinal information. Chen \emph{et al.} \cite{chen2017using} further utilized the ordinal information and presented a Ranking-CNN, which contained a series of basic CNNs trained with ordinal age labels. The deep regression forests (DRFs) model was proposed in \cite{shen2018deep} to deal with  the heterogeneous age data. Pan \emph{et al.} \cite{pan2018mean} presented the mean-variance loss to learn a good age distribution. Li \emph{et al.} \cite{li2019bridgenet} proposed the BridgeNet with a novel bridge-tree structure to mine the continuous relation among age labels. Tan \emph{et al.} \cite{tan2019deeply} presented a deep hybrid-aligned architecture to capture multiple types of features with complementary information. Some researchers formulated age estimation as a label distribution learning problem \cite{gao2017deep,shen2017label} and achieved promising results. However, these methods learn a global estimator for all different persons and fail to model the personalized aging process.

Some researchers have investigated the compact model and achieved excellent performance~\cite{zhang2019c3ae,yang2018ssr}.
C3AE~\cite{zhang2019c3ae} explored the limits of the compact model for facial age estimation, which possesses only 1/2000 parameters compared with VGGNet. SSR-Net~\cite{yang2018ssr} utilized a coarse-to-fine strategy and refined the results with multiple stages. A novel network structure was proposed with only 0.32 MB memory overhead. The goal of these methods is to obtain as small a model as possible without significantly degrading performance. These methods are beyond the scope of this paper, as we still focus on further advancing the performance of age estimation.

\subsection{Meta Learning}
The reason why humans can learn from very few examples is that the learning process is usually based on the experience gained from other tasks. Likewise, meta-learning aims at training a model with a better capacity of learning new tasks \cite{lee2019meta}. Meta-learning is widely used in machine learning~\cite{munkhdalai2017meta,li2021meta}, especially few-shot learning \cite{snell2017prototypical}. Meta-learning methods can be divided into 3 categories~\cite{lee2019meta}: metric-based methods, model-based methods, and optimization-based methods.

Metric-based methods usually aim to learn an efficient distance function for similarity.
Vinyals \emph{et al.} \cite{vinyals2016matching} proposed the matching network to calculate the similarity between the test sample and support set samples. The weighting sum of the support set labels was treated as the predicted label.
Prototypical network \cite{snell2017prototypical} encoded inputs into one-dimension vectors and the similarity was defined as the distance between those vectors. Sung \emph{et al.}~\cite{sung2018learning} proposed the Relation Network for few-shot learning, which learns to learn a deep distance metric to compare a small number of samples within episodes.
Model-based methods use extra models to predict parameters of the network which is used to solve the actual problem \cite{jia2016dynamic}. Meta Networks \cite{munkhdalai2017meta} combined fast weight layers and slow weight layers for fast generalization to different tasks. Optimization-based methods customize the optimizing process to make the models generalize to different tasks \cite{andrychowicz2016learning}. Finn \emph{et al.} \cite{finn2017model} proposed an optimization algorithm MAML, which considers the losses across different tasks when updating parameters. In the end, the model trained with the MAML algorithm can be easily fine-tuned on new tasks. Ravi \emph{et al.} \cite{ravi2016optimization} cast the design of an optimization algorithm as a learning problem and proposed an LSTM-based meta-learner model to learn the optimization algorithm.
Encouraged by the success of meta-learning, we design a personalized estimator meta-learner, which learns to learn adaptive estimators for different people. Different from most existing meta-learning methods, our method takes auxiliary task information as the input to handle the zero-shot issue.

\section{Proposed Approach}

\begin{figure*}[t]
\begin{center}
   \includegraphics[width=1.0\linewidth]{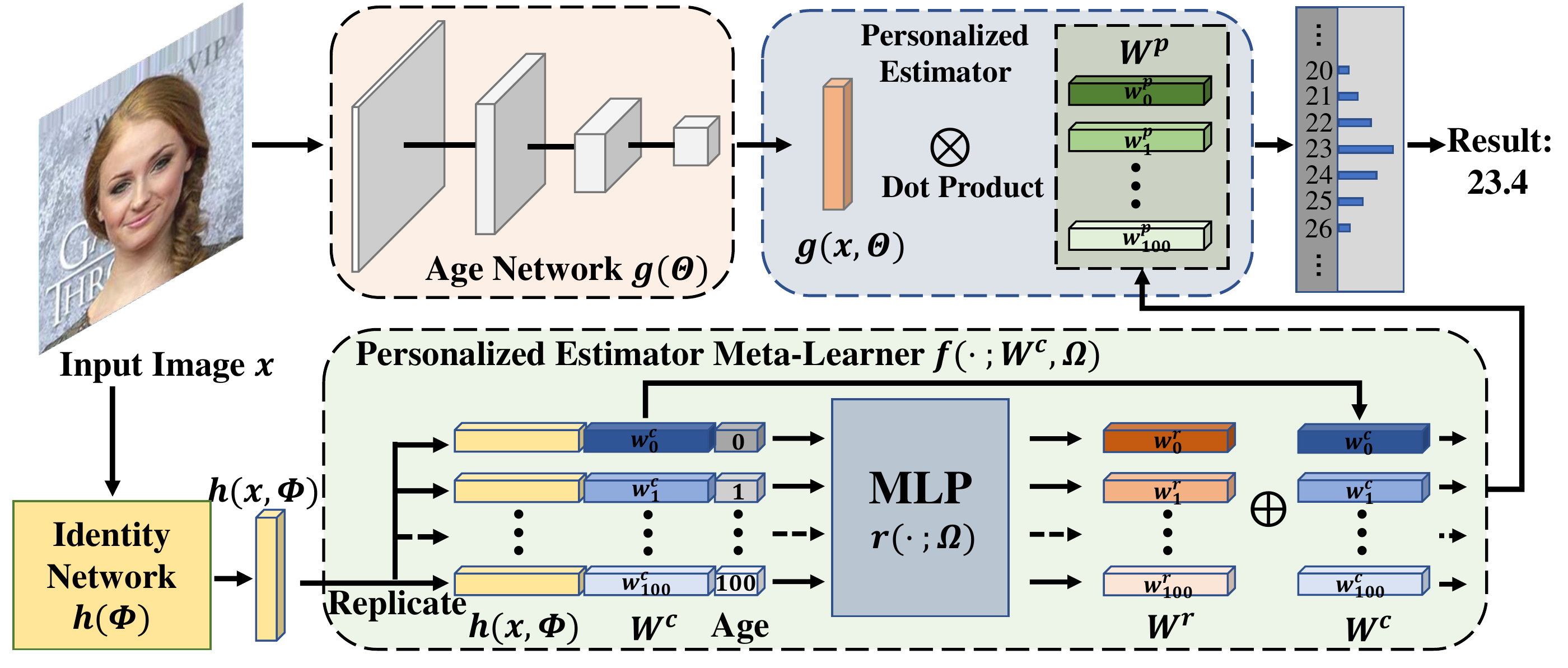}
\end{center}
   \caption{The overview of our proposed MetaAge. For an input image $\bm{x}$, we first send it to an age network $g(\bm{\Theta})$ to obtain the age features $g(\bm{x},\bm{\Theta})$. Meanwhile, the image $\bm{x}$ is also passed through an identity network $h(\bm{\Phi})$ to get the identity features $h(\bm{x},\bm{\Phi})$. Then our personalized estimator meta-learner generates the set of parameters $\{ \bm{w}_0^{p}, \bm{w}_1^{p}, ..., \bm{w}_{K-1}^{p} \}$ with different age inputs following \eqref{equ:residual}. The estimated age is calculated with age features $g(\bm{x},\bm{\Theta})$ and the customized estimator parameterized by $\bm{W}^{p}$ according to \eqref{equ:commonscore} - \eqref{equ:expectation}.}
\label{fig:flowchart}
\end{figure*}

In this section, we first review the formulations of global age estimators. Then we present the ideas of our method and provide an in-depth analysis of how the proposed personalized estimator meta-learner transfers the learned meta knowledge to unseen persons. Lastly, we introduce the design details of the proposed MetaAge. Fig. \ref{fig:flowchart} depicts an overview of our proposed approach.

\subsection{Learning Global Age Estimators}
We start with a brief introduction to a global estimator $\bm{e}$. Let $\bm{x}$ denotes an input sample and $y \in \{0, 1, ..., K-1\}$ denotes the corresponding age label. We consider implementing  the estimator $\bm{e}$ based on classification as in \cite{rothe2018deep}, where a $K$-way classifier is learned by treating age labels as independent classes. The input sample $\bm{x}$  is usually sent to a CNN $g(\cdot)$ parameterized by  $\bm{\Theta}$ to extract age features: $g(\bm{x}, \bm{\Theta}) \in \mathbb{R}^D$, where $D$ is the feature dimension. Then the estimator $\bm{e}$ is implemented by a fully connected layer parameterized by the weight $\bm{W} \in \mathbb{R}^{K \times D}$ and the bias $\bm{b} \in \mathbb{R}^{K}$. We rewrite $\bm{W}$ and $\bm{b}$  as $[\bm{w}_0, \bm{w}_1, ..., \bm{w}_{K-1}]^T$ and  $[b_0, b_1, ..., b_{K-1}]^T$ respectively, where $\cdot ^T$ denotes transposition. Thus, the class score for age $i \in \{0, 1, ..., K-1\}$ is formulated as:
\begin{equation}
s_i(\bm{x}) =  \bm{w_i}^T g(\bm{x},\bm{\Theta}) + b_i.
\label{equ:commonscorewithb}
\end{equation}

Here we zero the bias term ($\bm{b} = \bm{0}$) so that the score function is only parameterized by the weight $\bm{W}$:
\begin{equation}
s_i(\bm{x}) =  \bm{w}_i^T g(\bm{x},\bm{\Theta}).
\label{equ:commonscore}
\end{equation}

Then we have the probability distribution of ages by  using the softmax function:
\begin{equation}
p_i(\bm{x}) = \frac{\exp(s_i(\bm{x}))}{\sum_{k=0}^{K-1} \exp(s_k(\bm{x}))},
\label{equ:softmax}
\end{equation}
where $p_i(\bm{x})$ represents the probability that the age of input sample $\bm{x}$ is $i$.
As suggested in \cite{rothe2018deep}, the final age $\hat{y}(\bm{x})$ is  estimated by calculating the expectation of the above
 probability distribution:
\begin{equation}
\hat{y}(\bm{x}) =\sum_{k=0}^{K-1} k * p_k(\bm{x}).
\label{equ:expectation}
\end{equation}

As we can see, the learned estimator $\bm{e}$ is parameterized by $\bm{W}$, which is the same for all different people once learned. Since different people age in different ways, learning personalized age estimators for different people can better model the personalized aging processes.

\subsection{Learning to Learn Personalized Age Estimators}

Now we consider how a personalized age estimator $\bm{e}$ can be obtained with any available large-scale dataset. We first assume that we have the identity labels to show the issue of insufficient samples per person can be addressed by meta-learning. We reorganize the training set and test set into $\{\mathcal{D}_1, \mathcal{D}_2, ..., \mathcal{D}_n\}$ and $\{\mathcal{D}_{n+1}, \mathcal{D}_{n+2}, ..., \mathcal{D}_{n+m}\}$ according to the identity labels, where $n$ and $m$ represent the number of identities in the training set and the test set respectively, and $\mathcal{D}_j (1 \leq j \leq n+m)$ denotes a set of all samples of a person in the training/test set. Each set $\mathcal{D}_j$ corresponds to a task $\mathcal{T}_j$, whose objective is to learn a personalized age estimator on the set $\mathcal{D}_j$. Most existing personalized methods directly learn the parameters of a personalized estimator for each set $\mathcal{D}_j$ as illustrated in Fig. \ref{fig:introduction}. 
However, most sets $\mathcal{D}_j$ are relatively small, given that there are only a few samples for each person in the existing age datasets.
Therefore, it's infeasible to directly learn the parameters of an estimator on a set $\mathcal{D}_j$ with deep learning based methods.

Although most sets $\mathcal{D}_j$ only contain a few samples, we do have many sets $\{\mathcal{D}_1, \mathcal{D}_2, ..., \mathcal{D}_n\}$ for training, which correspond to many tasks $\{\mathcal{T}_1, \mathcal{T}_2, ..., \mathcal{T}_n\}$. Inspired by the success of meta-learning in the field of few-shot learning~\cite{finn2017model}, we consider learning to learn personalized estimators rather than directly learning the parameters of estimators. Humans can learn new skills and adapt to unseen situations rapidly. To empower the current AI systems with this ability, we need them to learn how to learn new tasks faster. Meta-learning systems usually use a large number of training tasks to learn how to adapt to new tasks. With the above formulated tasks, we propose a personalized estimator meta-learner, which learns to learn personalized age estimators. We train the personalized estimator meta-learner with tasks $\{\mathcal{T}_1, \mathcal{T}_2, ..., \mathcal{T}_n\}$ and test its ability on a new task $\mathcal{T}_{n + l} (1 \leq l \leq m)$, which is associated with the set $\mathcal{D}_{n+l}$.

Many meta-learning methods have been proposed in recent years, such as MAML \cite{finn2017model}, Prototypical Networks \cite{snell2017prototypical}, and MANN \cite{santoro2016meta}. However, directly applying the above meta-learning methods is not very suitable for personalized age estimation. The reason is that most of them usually require a few labeled samples on the test tasks, which is unrealistic considering that we cannot access the age labels on the test set. In general, we expect an age estimation method to work not only for the people in the training set but also for the people who have never been seen before. It means that we can find a set $\mathcal{D}_{n+l'} (1 \leq l' \leq m)$ in the test dataset whose identity does not exist in the training dataset. Then the corresponding task $\mathcal{T}_{n+l'}$ is a completely new task, and no labeled samples are available for this task. Therefore, we are confronted with the scenario of zero-shot learning. It is known that, to solve the zero-shot learning problem and transfer the learned knowledge to unseen persons, some auxiliary information that can represent the semantic relations among different tasks (identities) is necessary~\cite{wang2019survey}. We now end the assumption of the availability of identity labels and denote the auxiliary information as $\bm{I}$, which is identity information. Furthermore, our MetaAge proposes a meta-learner that takes identity information $\bm{I}$ as input and directly outputs the parameters of the corresponding age estimator.

\begin{figure*}[t]
  \centering
  \includegraphics[width=1.0\linewidth]{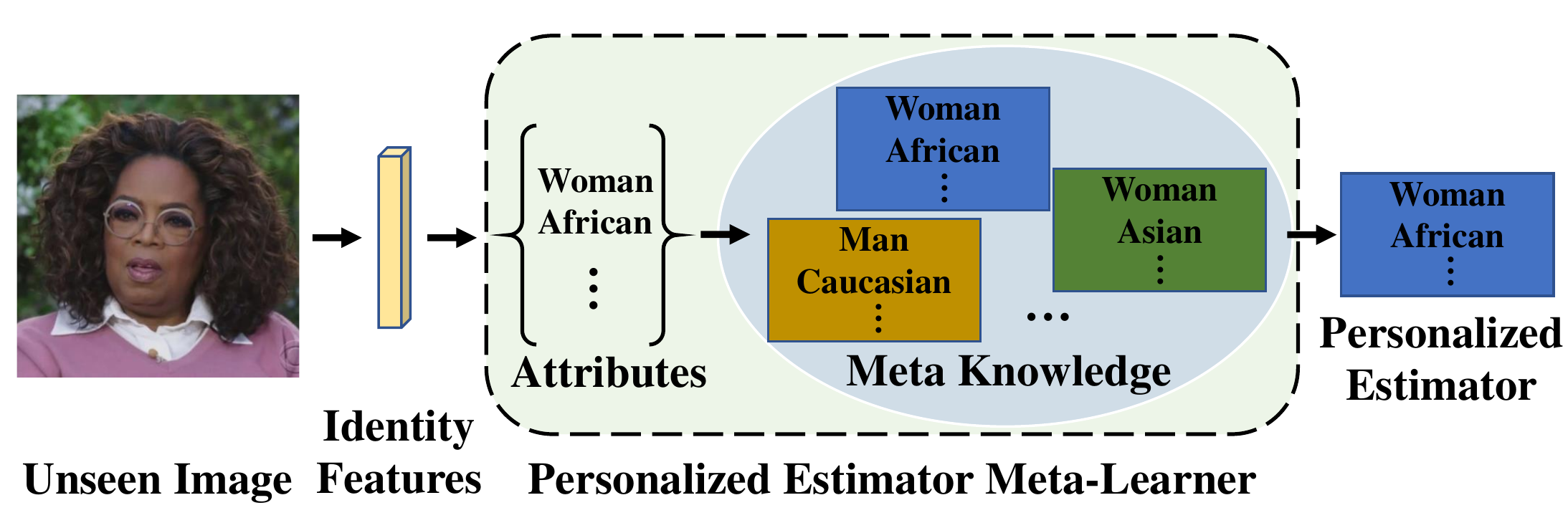}
  \caption{One way to understand our method. Our MetaAge learns the knowledge of  how identity-related attributes affect the parameters of personalized estimators. For an unseen person, our method transfers the meta knowledge based on the extracted attribute information and produces an accurate personalized age estimator. Note that this is for illustration only and we do NOT explicitly learn an estimator for each attribute.}
  \label{fig:understand}
\end{figure*}

Mathematically, we formulate the personalized estimator meta-learner with a one-step assignment operation conditioning on the identity information $\bm{I}$:
\begin{equation}
\bm{W}^{p} = f(\bm{I},\bm{\Omega}),
\label{equ:metalearner}
\end{equation}
where $f(\cdot)$ represents the proposed personalized estimator meta-learner parameterized by $\bm{\Omega}$, and $\bm{W}^{p}$ is the learned parameters of a personalized age estimator $\bm{e}$ for the person with identity information $\bm{I}$. The function $f(\cdot)$ is implemented by a neural network and learns how to learn adaptive age estimators based on the identity information $\bm{I}$. Once learned, the meta-learner generates the parameters $\bm{W}^{p}$ based on the identity information for any given test task. Then the personalized estimator parameterized by $\bm{W}^{p}$ is used for age estimation according to \eqref{equ:commonscore} - \eqref{equ:expectation}.

The remaining issue is how to attain the identity information $\bm{I}$. A natural choice is to use identity features extracted from a well-trained face recognition model. In fact, identity features have been widely used to represent identity information in many tasks, such as face clustering~\cite{yang2020learning} and face aging~\cite{liu2019attribute}. We also adopt the identity features as the identity information $\bm{I}$ given that they provide meaningful and  unified semantic representations. Besides, we can easily obtain the identity features due to the availability of well-trained face recognition networks. Formally, we use $h(\bm{\Phi})$ to represent the well-trained face recognition model, where $\bm{\Phi}$ is the parameter of function $h(\cdot)$. We reformulate the proposed MetaAge with identity features $h(\bm{\Phi})$ as follows:
\begin{equation}
\bm{W}^{p} = f(h(\bm{\Phi}),\bm{\Omega}).
\label{equ:AEML}
\end{equation}

\textbf{How to understand the MetaAge?} We provide a way to understand how the proposed personalized estimator meta-learner generates an accurate adaptive estimator even for an unseen person. Liu \emph{et al.} \cite{liu2015deep} conducted experiments with the output of the FC layer of a face recognition model and found that humans can easily assign each neuron in the identity features with a semantic concept it measures. They also observed that most of these concepts were intrinsic to face identities, such as gender, race, and the shape of facial components. Therefore, identity features contain rich identity-related attribute information. We also conducted our experiments to validate this claim and provided the results in the following section.

We implement the proposed meta-learner by a neural network, which takes the identity features $h(\bm{\Phi})$ as input and outputs the estimator parameters $\bm{W}^{p}$. Therefore, the MetaAge learns the mapping from identity-related attribute information to the parameters of personalized estimators. In other words, during the training phase, the MetaAge learns the knowledge of how identity-related attributes affect the parameters of personalized estimators. For example, the MetaAge learns the effect of different races on personalized estimators. Then the MetaAge needs to transfer the learned knowledge to a test task. We may never see the person on the test task, but we can have the information about the person's attributes through the extracted identity features. The learned knowledge of how identity-related attributes affect estimators is transferred to the specific case based on the extracted attribute information. Note that the identity features provide a unified semantic representation on the training set and test set, and become the bridge for knowledge transfer.

We illustrate the above analysis in Fig. \ref{fig:understand}. The personalized estimator meta-learner has learned how identity-related attributes, such as gender and race influence the parameters of personalized estimators. For a given image of an unseen person, the corresponding identity features encode the attribution information, for example, an African woman. Then the meta knowledge is transferred based on the attribution information and our meta-learner generates the parameters of the corresponding personalized age estimator. In the end, an accurate personalized age estimator can be achieved.

\subsection{Personalized Estimator Meta-Learner}

\begin{algorithm}[t]
	\DontPrintSemicolon
	\KwIn{Training samples, number of ages $K$, pre-trained parameters $\bm{\Phi}$ of face recognition network $h(\cdot)$, iteration numbers $N$, and hyper-parameters $\lambda, \delta$.\;}
	\KwOut{Parameters $\bm{W^c} = [ \bm{w}_0^c,\bm{w}_1^c...,\bm{w}_{K-1}^c ]^T$, parameters $\bm{\Theta}$ of the network $g(\cdot)$, parameters $\bm{\Omega}$ of the network $r(\cdot)$.}
	\SetAlgoLined	
    Initialize $h(\cdot)$ with the pre-trained weights $\bm{\Phi}$.\;
    \For{$iter = 1,2,...,N$ }
     {
      Sample mini-batch of $b$ training images.\;
      Extract age features and identity features with $g(\cdot)$ and $h(\cdot)$ respectively.\;
      Compute the parameters $\bm{W}^{p}$ for each sample $\bm{x}$ in the mini-batch with \eqref{equ:residual} based on $h(\bm{x},\bm{\Phi})$.\;
      Calculate the class scores for each sample using \eqref{equ:commonscore}.\;
      Compute the mini-batch loss $\mathcal{L}^{total}$ with \eqref{equ:loss1}, \eqref{equ:loss2}, and \eqref{equ:loss3}.\;
      Update the parameters $\bm{W^c}$, $\bm{\Theta}$, and $\bm{\Omega}$ by descending the stochastic gradient: $\nabla \mathcal{L}^{total}$.\;
     }
\textbf{Return:} The parameters $\{\bm{W^c}, \bm{\Theta}, \bm{\Omega}\}$.
\caption{ The training procedure of our MetaAge} \label{alg:pattrain}	
\end{algorithm}

The formulation in \eqref{equ:AEML} gives a general framework of how to use a meta-learner $f(\cdot)$ to generate adaptive estimators for different people. Now we consider an instantiation corresponding to \eqref{equ:commonscore}. For an input sample $\bm{x}$, we can obtain the identity features $h(\bm{x},\bm{\Phi}) \in \mathbb{R}^F$, where $F$ denotes the dimension of identity features. Then the identity features $h(\bm{x},\bm{\Phi})$ are sent to the proposed meta-learner, which is implemented by a neural network. The meta-learner outputs the parameters $\bm{W}^{p}  \in \mathbb{R}^{K \times D}$, which are used to predict the age of $\bm{x}$ following \eqref{equ:commonscore} - \eqref{equ:expectation}. However, this results in a $K \times D$ dimensional output space, which is too large to be acceptable.
To address this issue, we rewrite $\bm{W}^{p}$ as $[\bm{w}_0^{p}, \bm{w}_1^{p}, ..., \bm{w}_{K-1}^{p}]^T$. Instead of outputting the entire parameters $\bm{W}^{p}$, we design a neural network $f(\cdot)$ to output the parameters $\bm{w}_{i}^{p} \in \mathbb{R}^{D} (0 \leq i \leq K-1)$. In other words, our network does not output the entire parameter matrix $\bm{W}^{p}$ but outputs a $D$-dimensional parameter vector, which greatly reduces the dimension of output space. The MetaAge is formulated as follows:
\begin{equation}
\bm{w}_i^{p} = f(h(\bm{x},\bm{\Phi}),i,\bm{\Omega}), 0 \leq i \leq K-1.
\label{equ:smallAEML}
\end{equation}
Note that we input the identity features $h(\bm{x},\bm{\Phi})$ and age $i$ to the network together and output the parameters $\bm{w}_i^{p}$. Considering that function $f(\cdot)$ generates the class weight $\bm{w}_i^{p}$ for class $i$, it should take $i$ as the input condition. In practice, the identity features $h(\bm{x},\bm{\Phi})$ and the class value $i$ are concatenated and sent to the network. To obtain the entire parameter matrix $\bm{W}^{p}$, we calculate the set of parameters $\{ \bm{w}_0^{p}, \bm{w}_1^{p}, ..., \bm{w}_{K-1}^{p} \}$ by repeating the above forward pass $K$ times with different class value inputs $\{0, 1, ..., K-1\}$.

To further reduce the learning difficulty and improve the training stability, we consider decomposing the parameters $\bm{w}_i^{p} \in \mathbb{R}^{D}$ into common parameters  $\bm{w}_i^c \in \mathbb{R}^{D}$ and an adaptive parameters-residual $\bm{w}_i^r \in \mathbb{R}^{D}$. We denote $\bm{W}^{p}$, $\bm{W}^c$, and $\bm{W}^r$ as $[\bm{w}_0^p, \bm{w}_1^p, ..., \bm{w}_{K-1}^p]^T$, $[\bm{w}_0^c, \bm{w}_1^c, ..., \bm{w}_{K-1}^c]^T$, and $[\bm{w}_0^r, \bm{w}_1^r, ..., \bm{w}_{K-1}^r]^T$, respectively. The common parameters $\bm{W}^c$ are utilized to model the shared common aging patterns for all people which are the same for different individuals, while the adaptive parameters-residual $\bm{W}^r$ is used to model the person-specific aging patterns which varies with different persons. That is to say, we let $\bm{W}^{p} = \bm{W}^c + \bm{W}^r$, where $\bm{W}^r$ is the function of identity features $h(\bm{x},\bm{\Phi})$ and $\bm{W}^c$ denotes additional learnable parameters.  We explicitly introduce the common parameters $\bm{W}^c$ to implement the meta-learner with residual strategy. Different from the formulation in \eqref{equ:smallAEML}, MetaAge with residual strategy is modeled as follows:
\begin{equation}
  \bm{w}_i^{p} = \bm{w}_i^c  + \bm{w}_i^r = f_r(h(\bm{x},\bm{\Phi}),i,\bm{W}^c,\bm{\Omega}), 0 \leq i \leq K-1,
  \label{equ:residual1}
\end{equation}
where $\bm{W}^c = [\bm{w}_0^c, \bm{w}_1^c, ..., \bm{w}_{K-1}^c]^T$ are learnable parameters and $f_r(\cdot)$ represents the proposed personalized estimator meta-learner with residual strategy. 
We can further expand \eqref{equ:residual1} as follows:
\begin{equation}
  \bm{w}_i^{p} = \bm{w}_i^c  + \bm{w}_i^r = \bm{w}_i^c + r(h(\bm{x},\bm{\Phi}),i,\bm{\Omega}), 0 \leq i \leq K-1,
  \label{equ:residual2}
\end{equation}
where function $r(\cdot)$ represents the adaptive parameters-residual $\bm{w}_i^r$. We use a multilayer perceptron (MLP) to implement the residual function $r(\cdot)$. It should be pointed out that if we let the residual function $r(\cdot)$ be $\bm{0}$, then our method degenerates into a global-based method, which essentially learns a global estimator parameterized by $\bm{W^c} = [\bm{w_0}^c,\bm{w_1}^c,...,\bm{w_{K-1}}^c]^T$. Once learned, the parameters $\bm{W^c}$ are the same for different individuals while the $\bm{W^r}$ is not.
Although the identity feature $h(\bm{x},\bm{\Phi})$ and class value $i$ are sufficient for generating the parameters-residual $\bm{w_i}^r$ as the conditional input of the neural network $r(\cdot)$, we find that it is beneficial to introduce the common parameter $\bm{w}_i^c$ to the conditional input.
Mathematically, we reformulate MetaAge with residual strategy as follows:
\begin{equation}
  \bm{w}_i^{p} = \bm{w}_i^c + r(h(\bm{x},\bm{\Phi}),\bm{w}_i^c,i,\bm{\Omega}), 0 \leq i \leq K-1.
  \label{equ:residual}
\end{equation}
Specifically, we concatenate the identity feature $h(\bm{x},\bm{\Phi})$,  common parameter $\bm{w}_i^c$, and class value $i$, and then send them to the network $r(\cdot)$ to obtain the adaptive parameters-residual $\bm{w_i}^r$. To attain the set of parameters $\{\bm{w_0}^r,\bm{w_1}^r,...,\bm{w_{K-1}}^r\}$, we query the neural network $r(\cdot)$ with different class values $i$ and corresponding common parameters $\bm{w}_i^c$ as conditional inputs. 
Finally, we obtain the parameters $\{\bm{w_0}^p,\bm{w_1}^p,...,\bm{w_{K-1}}^p\}$ with the residual strategy defined in \eqref{equ:residual}.

For a sample $\bm{x}$ with age label $y$, we obtain the parameters $\bm{W}^{p}$ with \eqref{equ:residual}. Then we predict the age of $\bm{x}$ according to  \eqref{equ:commonscore} - \eqref{equ:expectation} with the obtained parameters $\bm{W}^{p}$. The cross-entropy loss function is used to optimize our model:
\begin{equation}
\mathcal{L}^{cls}(\bm{x},y)  = - \log(\frac{\exp(s_y(\bm{x}))}{\sum_{k=0}^{K-1} \exp(s_k(\bm{x}))}).
\label{equ:loss1}
\end{equation}

The aging patterns are temporal data, which means the age labels are ordinal numbers. We can utilize the ordinal property to better guide the learning of MetaAge. The ordinal property means that for a 30-year-old person, we predict that he/she is more likely to be 40 (20) than 50 (10).
Then a hinge loss function $H(z,z') = \max(0,\delta - (z - z'))$, where $\delta$ denotes the margin and is a hyper-parameter,
is utilized to model the ordinal property:
\begin{equation}
\mathcal{L}^{ord}(\bm{x},y) = \sum_{k=0}^{y-1}H(s_{k+1}(\bm{x}),s_{k}(\bm{x})) + \sum_{k=y}^{K-2}H(s_{k}(\bm{x}),s_{k+1}(\bm{x})).
\label{equ:loss2}
\end{equation}

We adopt the joint supervision of the above two losses to train our model:
\begin{equation}
\mathcal{L}^{total}(\bm{x},y)  = \mathcal{L}^{cls}(\bm{x},y)  + \lambda \mathcal{L}^{ord}(\bm{x},y) ,
\label{equ:loss3}
\end{equation}
where the parameter $\lambda$ balances two loss functions. It should be noted that the identity network $h(\bm{x},\bm{\Phi})$ is \emph{only} used to extract identity features and its parameters $\bm{\Phi}$ are \emph{NOT} updated during training.

In this way, our method addresses both requirements of existing personalized methods for datasets, which enables us to use the existing large-scale datasets  without any additional annotations. In the end, the proposed personalized estimator meta-learner can be plugged into any deep neural network and trained end-to-end to fully utilize the advantage of large-scale datasets. To better understand our method, we present the training procedure in algorithm \ref{alg:pattrain}.

\section{EXPERIMENTS}
In this section, we conducted extensive experiments on the widely-used MORPH II \cite{ricanek2006morph},  ChaLearn LAP 2015 \cite{escalera2015chalearn}, and ChaLearn LAP 2016 \cite{escalera2016chalearn} databases to demonstrate the effectiveness of the proposed MetaAge.

\subsection{Datasets}

\textbf{MORPH II:} The MORPH II database \cite{ricanek2006morph} consists of 55,134 images from about 13,000 subjects and the age range lies from 16 to 77 years old. We adopt two popular protocols for MORPH II in this paper. Following \cite{chang2011ordinal,agustsson2017anchored,rothe2018deep}, only 5,492 images of Caucasian Descent people from 2,193 individuals are used to reduce the cross-ethnicity influence for the first protocol. Then we randomly select 80 percent images for training and the remaining 20 percent images for testing. The second protocol is employed in \cite{shen2017label,chen2017using}, which randomly splits all of the images in MORPH II into two parts for training and testing by a ratio of four to one.
Following the practice of previous works~\cite{pan2020self,shen2018deep}, we adopt five-fold cross-validation on both protocols of the MORPH II dataset.

\textbf{ChaLearn LAP 2015:} The ChaLearn LAP 2015 database \cite{escalera2015chalearn} was used for apparent age estimation, which includes 4,699 images with age ranges from 0 to 100 years.  The standard train/val/test split uses 2,476 images for training, 1,136 images for validation, and 1,087 images for testing. The images were labeled by at least 10 users and the average age was treated as the final annotation.

\textbf{ChaLearn LAP 2016:} The ChaLearn LAP 2016 database \cite{escalera2016chalearn} was employed for the second edition of the competition of apparent age estimation. This database has been extended to 7,591 images. All images were split into three subsets: 4,113 images for training, 1,500 images for validation, and 1,978 images for testing. Each image of this database was annotated with a mean age and a corresponding standard deviation, which were calculated based on at least 10 human voters per image leading to nearly 145,000 votes for the database.

\subsection{Evaluation Metrics}
For MORPH II datasets, we employ the mean absolute error (MAE) and cumulative score (CS). The MAE is computed as the average of the absolute errors between the estimated ages and the ground truth ages:
\begin{equation}
MAE =  \frac{1}{M} \sum_{m=1}^{M} |\hat{y}_m - y_m|,
\label{equ:MAE}
\end{equation}
where $y_m$ is the ground truth age for the $m^{th}$ test image, $\hat{y}_m$ is the corresponding estimated age, and $M$ denotes the total number of test images. The CS metric is defined as follows:
\begin{equation}
CS(\theta) = (M_{\theta} /  M) \times 100\%,
\label{equ:CS}
\end{equation}
where $M_{\theta}$ represents the number of test images that have the absolute prediction error no more than $\theta$ (years). For apparent age estimation, the $\epsilon$-error is adopted as the evaluation metric, which is computed a:
\begin{equation}
\epsilon = 1 - \frac{1}{M} \sum_{m=1}^M \exp (- \frac{(\bm{\hat{y}_m - y_m} )^2}{2 \sigma_m^2}),
\label{equ:error}
\end{equation}
where $\sigma_m$ is the standard variation of the annotations for the $m^{th}$ test sample.

\subsection{Implementation Details}
Following the previous method in \cite{li2019bridgenet}, we detected each face using a face detector MTCNN \cite{zhang2016joint} and performed face alignment.  All the aligned faces were resized and cropped into 224 $\times$ 224. Then these images were sent to $g(\bm{\Theta})$ and $h(\bm{\Phi})$ to extract age features and identity features respectively. As the most popular backbone \cite{li2019bridgenet,pan2018mean,shen2018deep,rothe2018deep} for age estimation, VGG-16 was utilized to implement $g(\bm{\Theta})$, which was pre-trained with the IMDB-WIKI database. The age features were obtained from the $D=4096$ dimensional outputs of the penultimate fully connected layer. For $h(\bm{\Phi})$, we employed the ResNet-50 version of VGGFace2 \cite{cao2018vggface2} to get the $F=2048$ dimensional identity features. The residual function $r(\bm{\Omega})$ was implemented by a two-layer MLP with batch normalization \cite{pmlr-v37-ioffe15}. The function $r(\bm{\Omega})$ took the concatenation of $h(\bm{x},\bm{\Phi})$, $\bm{w}_i^c$ and $i$ as the input. In the experiments, we found that one-hot encoding of the age input achieved better results, so we adopted it in the following experiments. Given that $K$ was 101, the input dimension of $r(\bm{\Omega})$ was 6245 = 4096 (the dimension of $\bm{w}_i^c$) + 2048 (the dimension of $h(\bm{x},\bm{\Phi})$) + 101 (the dimension of one-hot encoding of class labels). Then the input was processed by $r(\bm{\Omega})$, which consists of a hidden layer with 8192 nodes and an output layer with 4096 dimensions.

To improve performance and avoid overfitting, data augmentation was utilized in our experiments. For each training image, random horizontal flipping and random cropping were applied.
The networks were optimized by Adam optimizer \cite{kingma2015adam} with $\beta_1=0.9$ and $\beta_2 =0.999$. We used $\lambda = 0.2$, $\delta = 2$ in the following experiments because those two parameters performed well in most cases. Generally, the initial learning rate was $10^{-4}$. Following \cite{li2019bridgenet}, the learning rate was reset to $10^{-5}$ for the ChaLearn LAP 2015 and  ChaLearn LAP 2016 datasets, considering that they have a relatively small amount of data. We trained our model for 60 epochs using mini-batches of 64.
The PyTorch \cite{Paszke2019PyTorch} packages were used to construct our module throughout the experiments.

\begin{figure*}[t]
  \centering
  \subfigure[Protocol I]{
    \label{fig:cs:morph1}
    \includegraphics[width=0.475\linewidth]{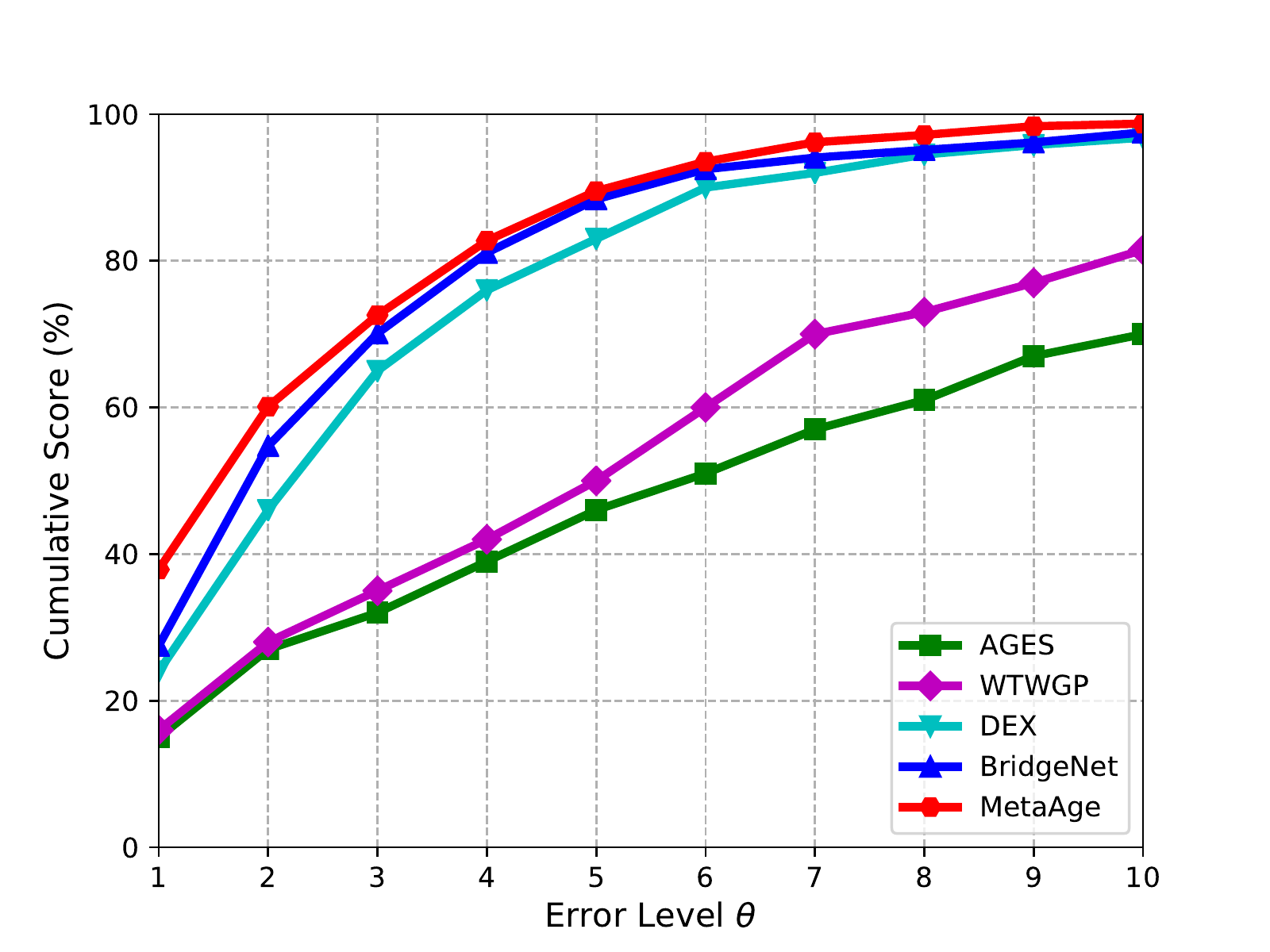}
  }
  \subfigure[Protocol II]{
    \label{fig:cs:morph2}
    \includegraphics[width=0.475\linewidth]{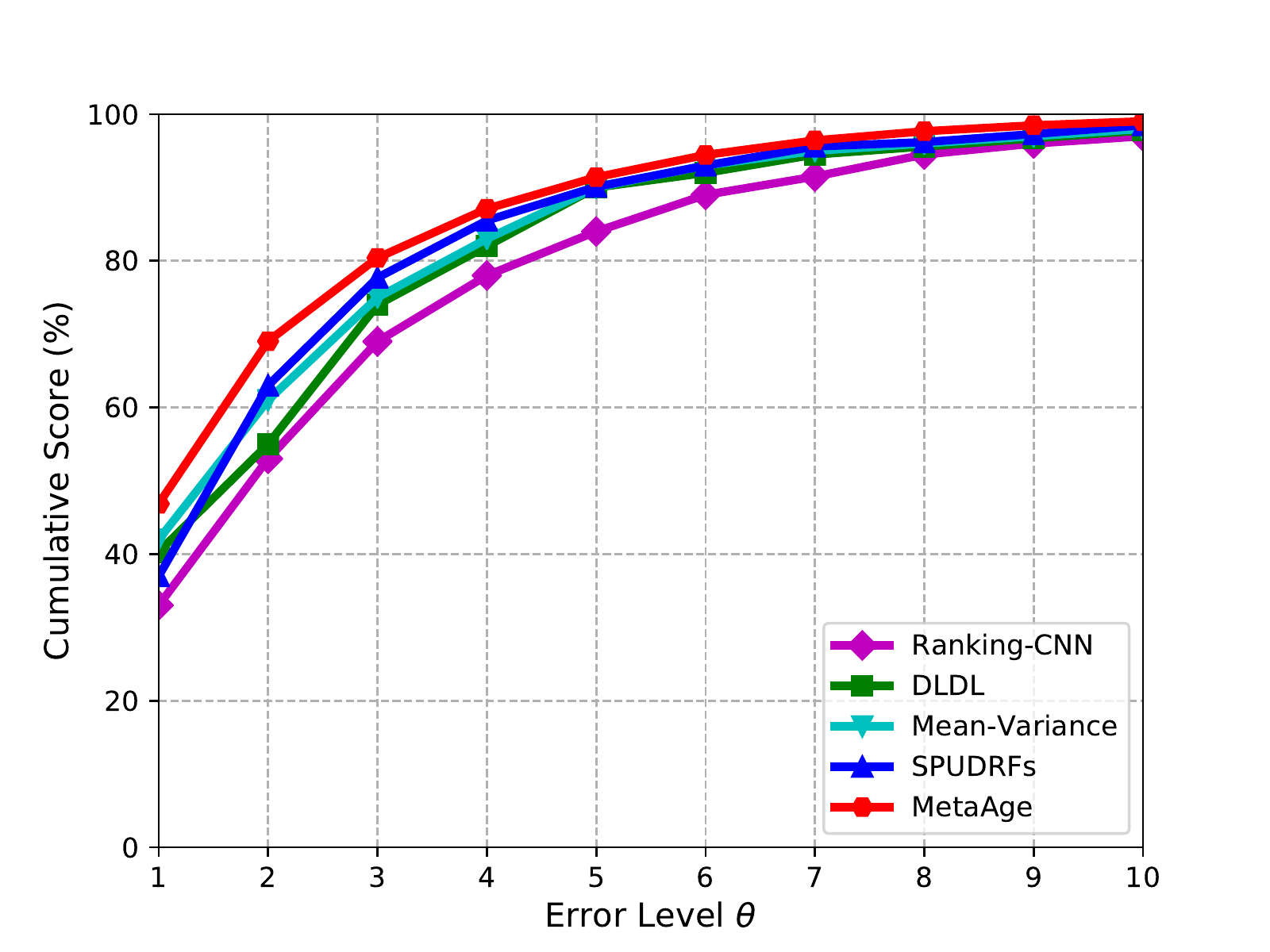}
  }
  \caption{The results of CS curves. (a) The comparisons with CS metric on the MORPH II dataset with
  protocol I. (b) The comparisons with CS metric on the MORPH II dataset with
  protocol II.}
  \label{fig:cs}
\end{figure*}

\begin{table}[t]
\caption{The comparisons between the proposed method and other state-of-the-art methods on the MORPH
II dataset. We report the MAE results under two different protocols. DL stands for deep learning based approach and PE means personalized age estimation method.}
\label{table:SOTA:MORPH}
\renewcommand\tabcolsep{5pt}
\centering
\begin{tabular}{lcccccc}
\toprule
Method  & Protocol I & Protocol II & DL & PE & Year\\
\midrule
AAS \cite{lanitis2004comparing} & 20.93 & - &  & & 2004\\
RUN   \cite{yan2007ranking} & 8.34 & -  &  & & 2007 \\
LARR  \cite{guo2008image} & 7.94 & - &  & & 2008\\
mkNN \cite{xiao2009learning} &  10.31  & - & & &  2009 \\
\midrule
AGES \cite{geng2007automatic} & 8.83 & - & & \Checkmark &2007 \\
MTWGP \cite{zhang2010multi} & 6.28 & - & & \Checkmark & 2010\\
\midrule
SSR-Net~\cite{yang2018ssr} & - & 2.52 & \Checkmark &  &  2018 \\
C3AE~\cite{zhang2019c3ae} & - & 2.75 &  \Checkmark &  &2019 \\
Ranking-CNN \cite{chen2017using} & - & 2.96 &  \Checkmark &  &2017 \\
DLDL \cite{gao2017deep} &  - & 2.42 &  \Checkmark &  &2017 \\
dLDLF \cite{shen2017label} & 3.02 & 2.24  &  \Checkmark &  &2017\\
DRFs \cite{shen2018deep} & 2.91 & 2.17 &   \Checkmark &  &2018\\
DEX \cite{rothe2018deep} & 2.68 & - &   \Checkmark &  &2018  \\
Mean-Variance \cite{pan2018mean} & - & 2.16 &   \Checkmark &  &2018 \\
DLDL-v2 \cite{gao2018age} & - & 1.97 &  \Checkmark &  &2018 \\
Tan \emph{et al.} \cite{tan2017efficient} & 2.52 & - & \Checkmark &  & 2018 \\
BridgeNet \cite{li2019bridgenet} & 2.38 & - &   \Checkmark &  &2019 \\
DHAA \cite{tan2019deeply} & 2.49 & 1.91 &  \Checkmark &  &2019 \\
AVDL \cite{wen2020adaptive} & 2.37 & 1.94 &  \Checkmark &  &2020 \\
SPUDRFs \cite{pan2020self} & - & 1.91 &  \Checkmark &  &2020 \\
\midrule
\textbf{MetaAge}  & \textbf{2.23} & \textbf{1.81} & \Checkmark & \Checkmark & - \\
\bottomrule
\end{tabular}
\vspace{-0.5cm}
\end{table}

\subsection{Comparisons with the State-of-the-Arts}

The MAE results on the MORPH II database are shown in Table \ref{table:SOTA:MORPH}, where DL stands for deep learning based approach and PE means personalized approach. The first six methods in Table \ref{table:SOTA:MORPH} are all non-deep learning methods, among which the first four are global-based methods and the last two are personalized methods. We observe that the personalized approaches generally outperform the global-based approaches. The reason is that personalized methods can better model the characteristics of personalized aging processes. However, these methods suffer from insufficient data of aging patterns, which severely limits the development of personalized methods. The remaining methods in Table \ref{table:SOTA:MORPH} are deep learning methods, which include state-of-the-art approaches. We see that deep learning methods outperform non-deep learning methods by a large margin because deep learning methods can learn a better feature representation with large-scale datasets and deep CNNs. The proposed MetaAge not only explicitly models the personalized aging patterns but also leverages existing data-driven deep learning techniques. As we can see, our method achieves the lowest MAE of 2.23 and 1.81 on MORPH II with the protocol I and protocol II respectively.  The proposed MetaAge significantly boosts the performance of previous personalized methods owing to the deeply learned features from the large-scale datasets. Since our method learns an adaptive age estimator for each individual, our approach also outperforms the state-of-the-art methods. To report the results of CS curves, we select state-of-the-art methods that reported CS curve results for comparison.
Fig. \ref{fig:cs} visualizes the CS curves on the MORPH II dataset under two protocols. We see that our proposed approach consistently outperforms other methods.

Two competition datasets of apparent age estimation were also employed to validate the proposed method. Following the tricks used in \cite{rothe2018deep,tan2017efficient,li2019bridgenet}, both training and validation sets were used to train our model in the training phase. To further improve the performance, we employed the 10-crop testing, which passed four crops from each corner and one crop from the center, as well as the horizontal flips of them through the networks. The final result was obtained by averaging these ten predictions. Since most ages in the ChaLearn LAP 2016 database were not integers and both databases provided the standard deviation $\sigma$ of annotations for each image, we employed the label distribution encoding of age labels instead of the one-hot encoding as the ground truths in the training stage following \cite{antipov2016apparent,liu2015agenet}.

\begin{table}[t]
\caption{Comparisons with the state-of-the-art methods on the ChaLearn LAP 2015 dataset.}
\label{table:rescha15}
\centering
\begin{tabular}{lcccc}
\toprule
Rank & Team Name & MAE & $\epsilon$-error & Single Model \\
\midrule
- & \textbf{MetaAge}  & \textbf{2.83} &  \textbf{0.250651} & \textbf{YES} \\
- & Tan \emph{et al.} \cite{tan2017efficient} & 2.94 & 0.263547 & NO \\
\midrule
1 & CVL\_ETHZ \cite{rothe2018deep}  & - & 0.264975   & NO \\
2 & ICT-VIPL \cite{liu2015agenet}  & - & 0.270685 & NO \\
3 & WVU\_CVL \cite{zhu2015study}   & - & 0.294835 &  NO  \\
4 & SEU\_NJU \cite{yang2015deep}  & - &  0.305763 & NO  \\
~ & Human & - & 0.34  & -  \\
5 & UMD &  - & 0.373352 & -  \\
6 & Enjuto  & - & 0.374390 & -\\
7 & Sungbin Choi & - & 0.420554 & - \\
8 & Lab219A & - & 0.499181 & -  \\
9 & Bogazici  & - & 0.524055 & - \\
10 & Notts CVLab & - & 0.594248 & -  \\
\bottomrule
\end{tabular}
\end{table}

\begin{table}[t]
\caption{Comparisons in $\epsilon$-error between our method and  the state-of-the-art methods on the ChaLearn LAP 2016 dataset.}
\label{table:rescha16}
\centering
\begin{tabular}{lcccc}
\toprule
Rank  & Team Name  & MAE & $\epsilon$-error & Single Model \\
\midrule
- & \textbf{MetaAge} & \textbf{3.49} & 0.2651 & \textbf{YES} \\
- &  Mean-Variance \cite{pan2018mean}& - & 0.2867 & YES \\
- &  Tan \emph{et al.} \cite{tan2017efficient} &  3.82 & 0.3100 & YES \\
\midrule
1 & OrangeLabs \cite{antipov2016apparent} & - & \textbf{0.2411} & NO \\
2 & palm\_seu \cite{huo2016deep} & - & 0.3214 & NO \\
3 & cmp+ETH & - & 0.3361 & NO \\
4 & WYU\_CVL & - & 0.3405 & NO \\
5 & ITU\_SiMiT \cite{can2016apparent} & - &  0.3668 & NO \\
6 & Bogazici \cite{gurpinar2016kernel} & - & 0.3740 & NO  \\
7 & MIPAL\_SNU & - & 0.4565 & NO \\
8 & DeepAge & - & 0.4573 & YES  \\
\bottomrule
\end{tabular}
\end{table}

The results on the ChaLearn LAP 2015 dataset are summarized in Table \ref{table:rescha15}. It is shown that our method achieves the best performance among all methods on the test set with an $\epsilon$-error of 0.250651. It should be noted that our method only uses one model, whereas other methods use an ensemble of multiple models. The comparisons between our method and the state-of-the-art methods on the ChaLearn LAP 2016 dataset are reported in Table \ref{table:rescha16}. We observe that our method is next only to OrangeLabs' method \cite{antipov2016apparent} and achieves an $\epsilon$-error of 0.2651. However, the OrangeLabs' method employs a private dataset and a manually cleaned IMDB-WIKI dataset. Moreover, an ensemble of 14 networks is utilized to further boost the performance of their method. Instead, our method only uses the publicly available datasets and a single model. Compared with the second-place method \cite{huo2016deep}, our method reduces the $\epsilon$-error by 0.0563 with a single model, which demonstrates the effectiveness of the proposed approach. Some state-of-the-art methods also report their results on this dataset with a single model and we see that our method achieves better performance, which illustrates the superiority of learning personalized age estimators.

\begin{table}[t]
  \caption{Cross-database evaluation on the FG-NET database (trained on the MORPH II database).}
  \label{table:crossdata}
  \renewcommand\tabcolsep{15pt}
  \centering
  \begin{tabular}{lccc}
  \toprule
  Method & DEX \cite{rothe2018deep} & DLDL \cite{gao2017deep} & MetaAge   \\
  \midrule
  MAE & 5.73 & 5.45 &  \textbf{5.25} \\
  \bottomrule
  \end{tabular}
\end{table}

\subsection{Cross-Database Evaluation}
The training and test sets of existing age estimation methods are usually derived from the same dataset.  However, the data in real scenarios often have different distributions and characteristics from the training dataset. To further evaluate the generalizability of the proposed method, we conducted experiments across datasets. Specifically, we train the model on one dataset and then test the performance on another dataset. This is a more challenging protocol, as the test dataset may have a completely different data distribution.

We use the training data of the MORPH II database (protocol II) as the training database and test the performance on the FG-NET database \cite{panis2016overview}. FG-NET database \cite{panis2016overview} has 1,002 facial images of 82 persons with large variations in pose, expression, and lighting. All the images from the FG-NET database are used for evaluation. For comparison, we also re-implemented two state-of-the-art methods and tested their performance in the cross-database setting. The results are presented in Table~\ref{table:crossdata}. We see that our proposed method provides the lowest MAE, which indicates that our method has better generalization.

\subsection{Ablation Study}
\textbf{Effect of Identity Features:} To transfer the learned meta knowledge to unseen persons, we introduced identity features to provide unified semantic representation. To validate that the superiority of MetaAge is not due to the introduction of identity features, we consider several different strategies to cooperate with identity information and conduct ablation experiments on the MORPH II dataset with the protocol I.

1) Fine-tuning. Instead of pre-training the age network $g(\bm{\Theta})$ on the IMDB-WIKI dataset, we use the  VGGFace~\cite{parkhi2015deep} pre-trained parameters to initialize the age network $g(\bm{\Theta})$, where VGGFace is a commonly used face recognition dataset. In this way, we encode the identity information in the initialized weights of $g(\bm{\Theta})$.

2) Learning without forgetting. Since the network may lose the ability of identification during fine-tuning, we consider a learning without forgetting strategy~\cite{li2017learningwf}. Concretely, we use a fixed face model as a teacher network $h_t(\bm{\Phi})$. We add an additional loss term to the age network $g(\bm{\Theta})$ to maintain its identification ability: $\mathcal{L}^{LwF} = |\cos(h_t(\bm{x}_1),h_t(\bm{x}_2)) - \cos(g(\bm{x}_1),g(\bm{x}_2)) |$, where $\cos(\cdot)$ denotes the cosine distance.

3) Multi-task learning. We explicitly introduce a face recognition task to exploit the identity information. Since no identity labels are available, we first use K-means to cluster faces based on identity features. Then we train two tasks jointly with the clustered pseudo labels and age labels.

4) Concatenating features. We first concatenate the identity features $h(\bm{x},\bm{\Phi})$ and age features $g(\bm{x},\bm{\Theta})$, and then learn a global estimator with the concatenated features. For a fair comparison, the global estimators were implemented with a two-layer MLP whose model size is similar to our method. Concretely, the global estimator consists of two hidden layers with 8192 dimensions and 4096 dimensions respectively, and a classification layer with 101 nodes.

\begin{table}[t]
\caption{Ablation studies of the identity features with different strategies on the MORPH II dataset (protocol I).}
\renewcommand\tabcolsep{25pt}
\label{table:identityfeature}
\centering
\begin{tabular}{lc}
\toprule
Methods  & MAE \\
\midrule
Baseline & 2.56  \\
\midrule
Fine-tuning & 2.62  \\
Learning without Forgetting & 2.49 \\
Multi-task Learning & 2.52 \\
Concatenating Features & 2.46 \\
\midrule
MetaAge & 2.23 \\
\bottomrule
\end{tabular}
\end{table}

Table \ref{table:identityfeature} shows the results. The Baseline in Table \ref{table:identityfeature} means learning a global estimator for age features $g(\bm{x},\bm{\Theta})$. We see that the Fine-tuning is even worse than the Baseline, which is reasonable since the weights pre-trained on the IMDB-WIKI dataset are proven to give better initialization~\cite{wen2020adaptive}. Both Learning without Forgetting and Multi-task Learning methods learn identity information from the supervision signals provided by the identity network. Meanwhile, the Concatenating Features solution directly uses the identity features extracted from the identity network, which better preserves the identity information. Therefore, the Concatenating Features strategy achieves the best performance among these three methods. Compared with the Baseline, the Concatenating Features strategy reduces MAE by 0.1 years. By contrast, our MetaAge outperforms the Baseline by 0.33 years, which is much significant. It demonstrates that the performance of our method mainly comes from the design of the proposed meta-learner rather than the introduction of identity features. Actually, the identity features are only used as the bridge to transfer the learned meta knowledge to unseen persons in our method. The superiority of MetaAge is mainly owing to the fact that our meta-learner generates adaptive estimators for different people while all the above methods still learn a global estimator.

\begin{table}[t]
\caption{More ablation results of the identity features.}
\label{table:more}
\centering
\begin{tabular}{l|c|cc}
\toprule
Metric & MAE & \multicolumn{2}{c}{$\epsilon$-error} \\
\midrule
\multirow{2}*{Database} & MORPH II  & \multirow{2}*{ChaLearn15} & \multirow{2}*{ChaLearn16} \\
~ & (Protocol II) & ~ &~ \\
\midrule
Baseline & 2.35 & 0.27287 & 0.3159 \\
Concatenating Features & 2.28 & 0.26455 & 0.3008 \\
MetaAge & \textbf{1.81} & \textbf{0.25065} & \textbf{0.2651} \\
\bottomrule
\end{tabular}
\end{table}

To further demonstrate that the performance of our method mainly comes from the design of the proposed meta-learner rather than the introduction of identity features, we provide more results on other databases in Table \ref{table:more}. We conducted experiments on the MORPH II database with protocol II,  ChaLearn LAP 2015 database, and ChaLearn LAP 2016 database. Since the Concatenating Features solution achieves the best performance among the four alternative strategies, we only compare our method with this solution in Table \ref{table:more}.
We observe that our MetaAge outperforms the Concatenating Features strategy by a large margin, which further illustrates the superiority of our method.

\begin{table}[t]
\caption{Ablation experiments of different components on the MORPH II dataset (protocol I).}
\label{table:ablation}
\centering
\begin{tabular}{l|ccc|ccc}
\hline
\multirow{2}*{Component} & \multicolumn{6}{c}{Age Network Backbone} \\
\cline{2-7}
~ & \multicolumn{3}{c|}{VGG-16} & \multicolumn{3}{c}{ResNet-50} \\
\hline \hline
Baseline & \Checkmark & \Checkmark & \Checkmark & \Checkmark  & \Checkmark &  \Checkmark \\
Meta-Learner & & \Checkmark  & \Checkmark &   &\Checkmark &  \Checkmark  \\
Residual Strategy& & &  \Checkmark  &   &   & \Checkmark \\
\hline
MAE & 2.56 & 2.34 & 2.23  & 2.62 &  2.30 & 2.17 \\
\hline
\end{tabular}
\end{table}

\begin{figure*}[t]
  \begin{center}
     \includegraphics[width=1.0\linewidth]{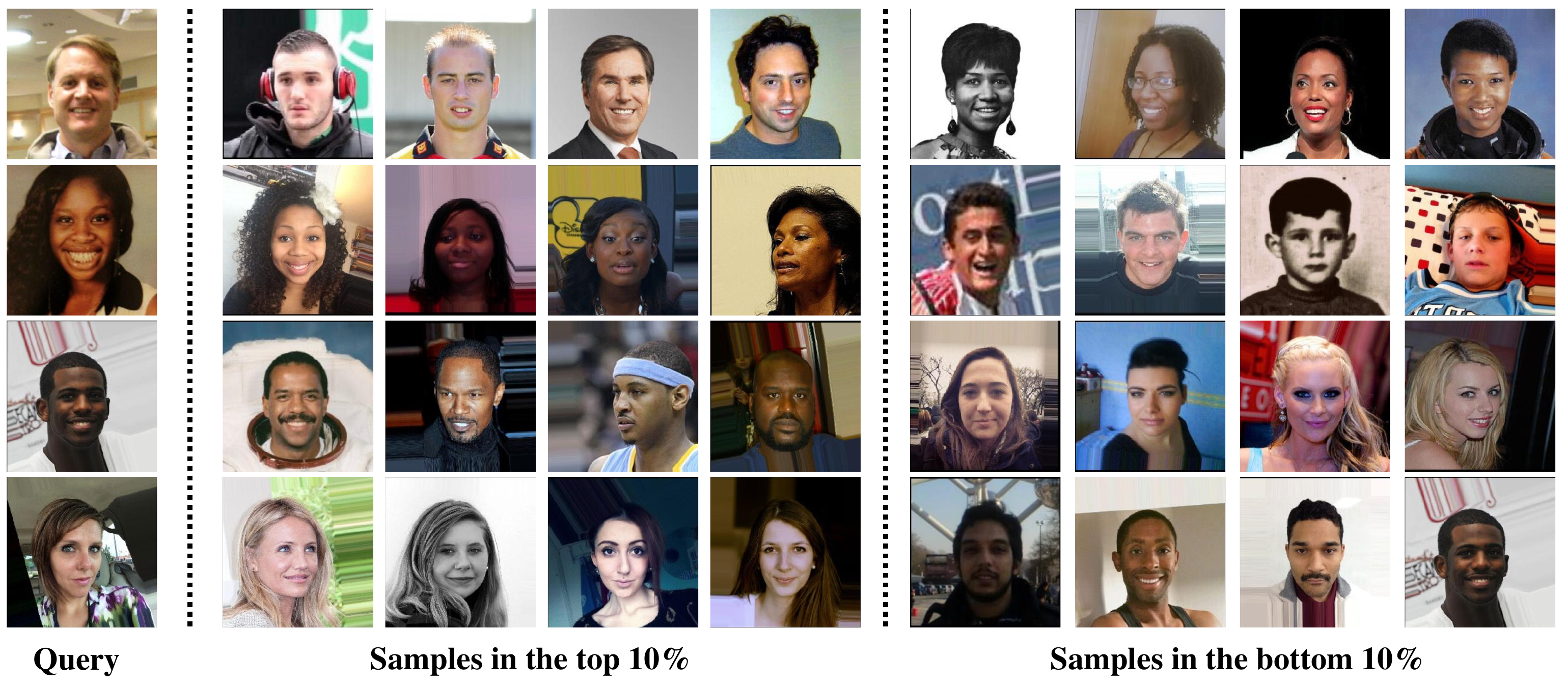}
  \end{center}
     \caption{Qualitative results. We utilize the parameters $\bm{W}^{p}$ as the features of query images and retrieved images. The retrieval results are obtained according to the Euclidean distance between the features of a query image and the features of retrieved images. Here we show some samples in the top 10\% and the bottom 10\%.}
  \label{fig:quali}
\end{figure*}

\textbf{Effect of Different Components:} We conducted ablation experiments to shows the influences of different components. Table \ref{table:ablation} shows the experimental results on the MORPH II dataset with protocol I. The Baseline means naive training age network $g(\bm{\Theta})$ while Meta-Learner represents training our method without the residual strategy. Compared with the baseline, the Meta-Learner solution improves performance by 0.22 years, which illustrates the effectiveness of learning to learn adaptive age estimators. We further observe that the residual strategy improves the performance to 2.23 years, which outperforms the baseline by 0.33 years for MAE. We also conducted experiments with different backbones and used ResNet-50 to implement age network $g(\bm{\Theta})$. We see that with ResNet-50 as the age network backbone, our method achieves an MAE of 2.17 years and outperforms the baseline by 0.45 years, which further illustrates the robustness of our method.

\begin{table}[t]
  \caption{Comparison results on six demographic groups. We report the MAE results on the MORPH II database under protocol II.}
  \centering
  \begin{tabular}{lccc}
  \toprule
  Demographic Group  & DEX~\cite{rothe2018deep}  & Mean-Variance~\cite{pan2018mean}  & MetaAge \\
  \midrule
  (women, african) & 2.19 & 2.17 & \textbf{2.07} \\
  (male, african) & 2.04 & 1.99 & \textbf{1.79} \\
  (women, caucasian) & 2.03 & 2.01 & \textbf{1.74} \\
  (male, caucasian) & 1.94 & 1.88 & \textbf{1.77} \\
  (women, asian) & 2.60 & 2.40 & \textbf{2.20} \\
  (male, asian) & 3.00 & 3.00 & \textbf{2.04} \\
  \bottomrule
  \end{tabular}
  \label{table:attributeour}
\end{table}
  
\begin{table}[t]
  \caption{Comparisons of previous attribute-based age estimation methods.}
  \label{table:attributeother}
  \renewcommand\tabcolsep{10pt}
  \centering
  \begin{tabular}{lccc}
  \toprule
  Metric & \multicolumn{2}{c}{MAE} & $\epsilon$-error \\
  \midrule
  \multirow{2}*{Database} & \multicolumn{2}{c}{MORPH II}  & \multirow{2}*{ChaLearn16} \\
  ~ & Protocol I & Protocol II &~ \\
  \midrule
  CMT~\cite{yoo2018deep} & - & 2.91 & -  \\
  RAGN~\cite{duan2017ensemble} & - & 2.61 & 0.3679 \\
  EGroupNet~\cite{duan2020egroupnet} & 2.48 & 2.13 & 0.3578 \\
  MetaAge & \textbf{2.23} & \textbf{1.81} & \textbf{0.2651} \\
  \bottomrule
  \end{tabular}
  \end{table}

\textbf{Comparisons with Attribute-based Methods:} To validate that our approach is personalized and not just attribute-based, we further analyze the results on different demographic groups. Since protocol II of the MORPH II database contains data from different ethnicities and the corresponding attribute labels, we adopt this setting for experiments. We first consider learning attribute-specific estimators with existing state-of-the-art methods~\cite{rothe2018deep,pan2018mean}. Specifically, we learn a model for each demographic group and select the corresponding model for prediction during the testing phase. The comparison results of all methods on six demographic groups are reported in Table \ref{table:attributeour}. Note that our method uses only one model while the other methods use six models to learn attribute-specific age estimators. The results show that our method is consistently superior to these attribute-based methods, which illustrates that our method can generate personalized age estimators based on  fine-grained identity information.

In addition, we also compare our approach with previous attribute-based methods. Table \ref{table:attributeother} shows the results. CMT~\cite{yoo2018deep} proposed to learn a gender-conditioned age probability with conditional multitask learning. RAGN~\cite{duan2017ensemble} included three convolutional neural networks: Age-Net, Gender-Net, and Race-Net, which explicitly uses gender and race information for age estimation. EGroupNet~\cite{duan2020egroupnet} utilized a feature-enhanced network to leverage age-related attributes including gender, race, hair, and expression. We see that our method significantly outperforms these methods, illustrating that our method exploits information beyond human attributes.

\begin{table}[t]
\caption{Ablation study of the global parameter (GP) in the residual strategy. We report the MAE results on the MORPH II database under protocol I.}
\renewcommand\tabcolsep{10pt}
\centering
\begin{tabular}{lccc}
\toprule
Age Network Backbone  &  Eq. \eqref{equ:smallAEML}  & Eq. \eqref{equ:residual2} & Eq. \eqref{equ:residual}  \\
\midrule
VGG-16 & 2.34 &  2.26 & 2.23   \\
ResNet-50  & 2.30 & 2.27 & 2.17 \\
\bottomrule
\end{tabular}
\label{table:gp}
\end{table}

\textbf{Ablation Study of the Global Parameter in the Residual Strategy:}
Our MetaAge exploits a residual strategy as shown in Eq. \eqref{equ:residual}. Compared with Eq. \eqref{equ:residual2}, the input of $r()$ includes the extra global parameter $\bm{w}^c_i$.  We provide the ablation study of this design choice on the MORPH II dataset under protocol I with different age network backbones in Table \ref{table:gp}.  We observe that the extra global parameters $\bm{w}^c_i$ are beneficial to our residual strategy, which is adopted in our experiments.

\subsection{Parameters Discussion}

In our paper, we set the $\lambda$ and $\delta$ to $0.2$ and $2$, respectively. Here we provide a detailed analysis of these parameters on the MORPH II dataset with protocol I.

We first set $\lambda = 0.2$ and only change the value of $\delta$ in the parameter searching process. The experimental results are shown in Table \ref{table:delta}. We observe that our method is relatively insensitive to $\delta$. The best results are achieved when $\delta = 2$ and we adopt this setting in our experiments.

\begin{table}[t]
  \caption{MAE results with a variety of $\delta$ on the MORPH II dataset with the protocol I.}
  \label{table:delta}
  \renewcommand\tabcolsep{5pt}
  \centering
  \begin{tabular}{l|cccccccc}
  \toprule
  $\delta$ & 0 & 0.1 & 1 & 2 & 3 & 4 & 5  & 10   \\
  \midrule
  MAE & 2.267 & 2.262 & 2.236 & \textbf{2.231} & 2.238 & 2.247 & 2.258 & 2.270 \\
  \bottomrule
  \end{tabular}
\end{table}

\begin{table}[t]
\caption{MAE results with a variety of $\lambda$ on the MORPH II dataset with the protocol I.}
\label{table:lambda}
\renewcommand\tabcolsep{5pt}
\centering
\begin{tabular}{l|cccccccc}
\toprule
$\lambda$ & 0 & 0.1  &  0.2 & 0.4 & 0.5 & 1 & 2 &10    \\
\midrule
MAE & 2.323 & 2.246 & \textbf{2.231} & 2.248 &  2.253 & 2.269 & 2.279 & 2.289 \\
\bottomrule
\end{tabular}
\end{table}

We further conducted experiments with different $\lambda$ ($\delta$ is set to 2). The experimental results are shown in Table \ref{table:lambda}. We observe that the use of $\mathcal{L}^{ord} (\lambda > 0)$ improves the performance of our method since $\mathcal{L}^{ord}$ explicitly models the ordinal property of age labels and provides complementary supervision for our model. The best results are achieved when $\lambda = 0.2$ and we adopt this setting in our experiments.

\begin{figure*}[t]
  \begin{center}
     \includegraphics[width=1.0\linewidth]{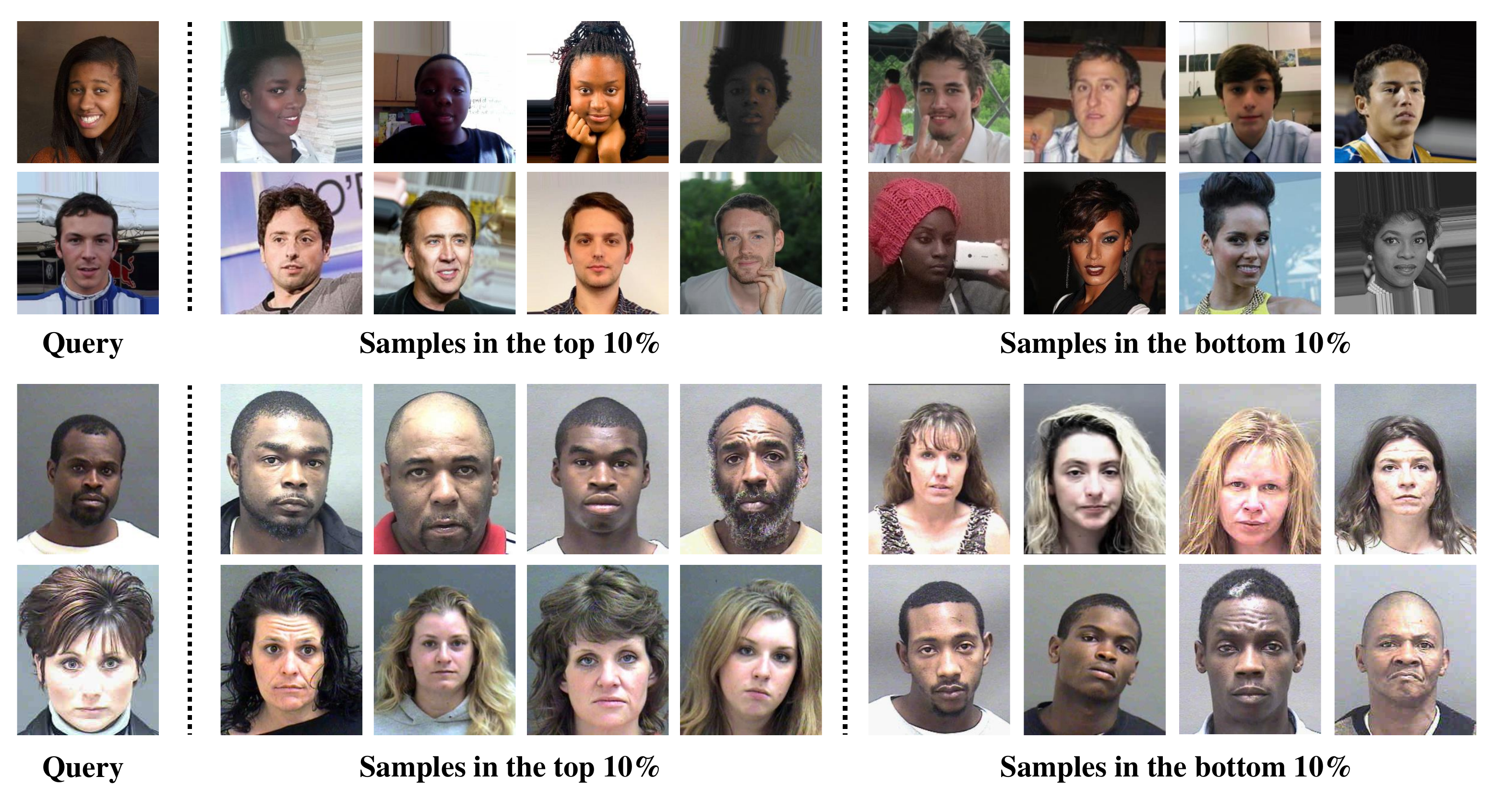}
  \end{center}
     \caption{More qualitative results on the ChaLearn LAP 2015 database and MORPH II database. The first two rows show the results on the ChaLearn LAP 2015 database, and the last two rows show the results on the MORPH II database.}
  \label{fig:morevisual}
  \end{figure*}

\subsection{Qualitative Evaluation}
To show that our method has learned how to generate a personalized age estimator based on identity information, we consider an image retrieval task for qualitative analysis.
For a facial image, we use the parameters $\bm{W}^{p}$ generated by our personalized estimator meta-learner as the retrieval features of this image. We sort the retrieved images according to the Euclidean distance between the retrieval features. Therefore, the corresponding estimators of the top-ranking retrieved images are similar to the estimator of the query image. We conducted experiments on the ChaLearn LAP 2016 dataset, where the train set was used for training the meta-learner and the test set was used for image retrieval. We randomly selected one image in the test set as the query image and set all the remaining test images as the retrieval images. Fig. \ref{fig:quali} shows the results and we observe that the samples in the top 10\% of the retrieval results have a higher identity similarity with the query image (they share more identity-related attributes, such as race and gender) than those in the bottom 10\%, which illustrates that the proposed MetaAge learned the knowledge of how to learn personalized estimators and could generate more similar estimators for samples with higher identity similarity. 
We further use the proxy task of image retrieval on the ChaLearn LAP 2015 and MORPH II databases for qualitative evaluation. We also use their training sets to train the meta-learner separately and perform image retrieval on the test set. For the MORPH II database, we adopt protocol II as it includes data from different races. We visualize the results in Fig. \ref{fig:morevisual}. We observe that higher identity similarity leads to more similar estimators, which validates that our method can generate personalized age estimators based on identity information. 

\begin{figure}[t]
  \begin{center}
     \includegraphics[width=1.0\linewidth]{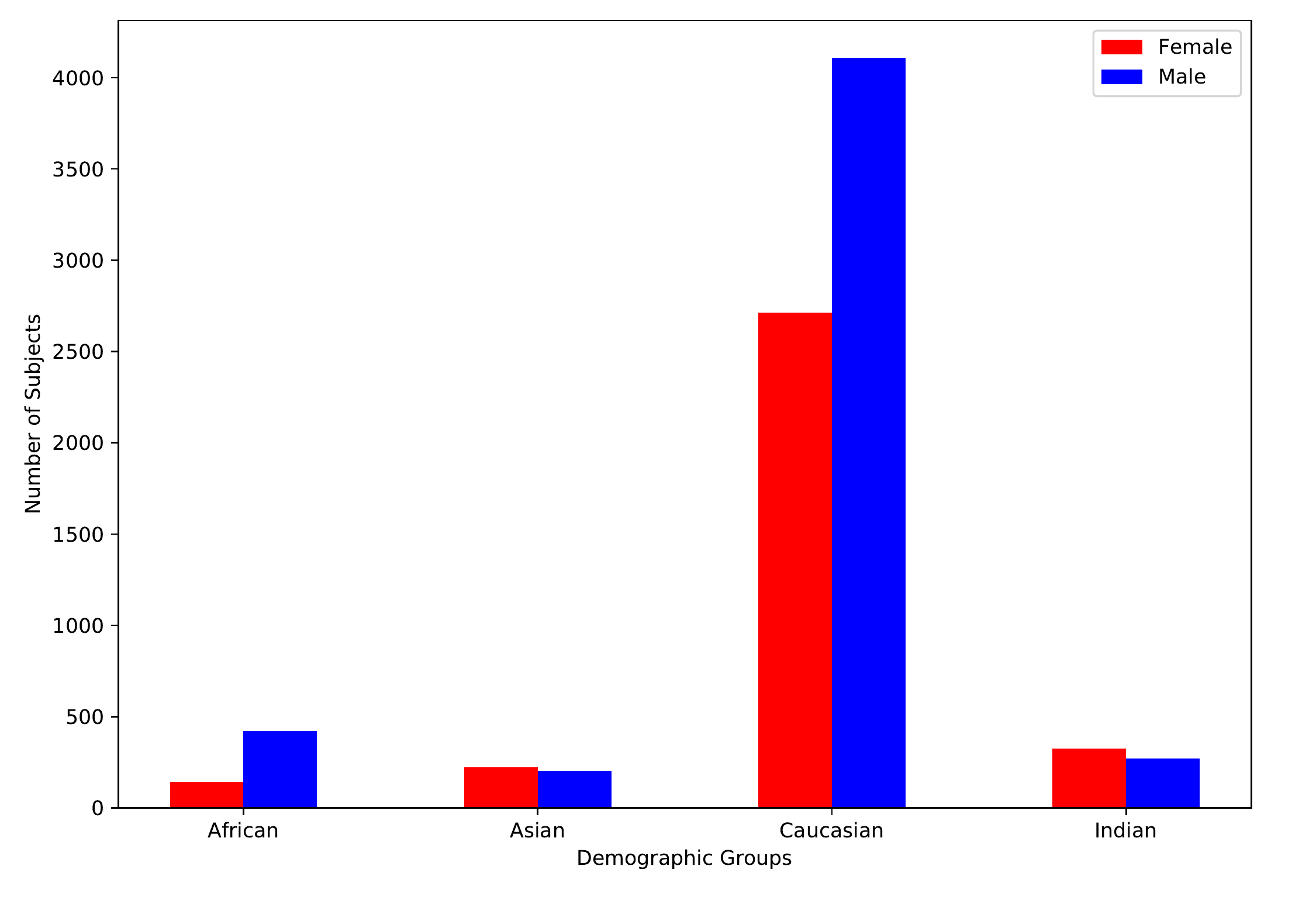}
  \end{center}
     \caption{The distribution of demographics on the VGGFace2 dataset. We see that there is a significant distribution imbalance between races.}
  \label{fig:vggface2}
\end{figure}

\subsection{Discussion}
To generate personalized age estimators, our approach utilizes identity features to provide identity information. One problem with using pre-trained face recognition models to extract identity features is that our method may inherit their biases. Since our approach uses the VGGFace2 pre-trained face recognition model, we present the distribution of demographics on the VGGFace2 database in Figure \ref{fig:vggface2}. We observe that the majority of individuals in this dataset are Caucasian. Using such an unbalanced dataset, our method naturally performs better for Caucasian populations, which is also verified by the results in Table \ref{table:attributeour}. To address this issue, we can use more balanced datasets to train face recognition networks or further develop unbiased face recognition algorithms.

\section{Conclusions}

In this paper, we have presented the MetaAge, which consists of a personalized estimator meta-learner to explicitly model the personalized aging processes. Instead of learning the parameters of an adaptive age estimator for each individual, as the most personalized methods did, our method learns the mapping from identity information to age estimator parameters.
The proposed MetaAge does not require the age datasets to contain identity labels and enough samples for each person,  which enables our approach to leverage any existing large-scale age estimation datasets without any additional annotations. Extensive experimental results on the MORPH II, ChaLearn LAP 2015, and ChaLearn LAP 2016 datasets demonstrate the effectiveness of our method.
The success of our approach sheds light on data-driven personalized age estimation methods and may also be meaningful for generic transfer learning tasks, which are interesting directions for our future work.

{
\bibliographystyle{IEEEtranS}
\bibliography{bibmetaage}
}

\end{document}